\documentclass[10pt, a4paper]{article}
\usepackage{lrec}
\usepackage{graphicx}
\usepackage{tabularx}
\usepackage{soul}
\usepackage{times}
\usepackage{latexsym}
\usepackage{url}
\usepackage{adjustbox}
\usepackage{amsmath}
\usepackage{epstopdf}
\usepackage[latin1]{inputenc}
\usepackage{hyperref}
\usepackage{xstring}

\title{Recommendation Chart of Domains for Cross-Domain Sentiment Analysis: Findings of A 20 Domain Study}

\name{Akash Sheoran\textsuperscript{$\diamondsuit$}, Diptesh Kanojia\textsuperscript{$\dagger$,$\clubsuit$,$\star$}, Aditya Joshi\textsuperscript{$\heartsuit$}, and Pushpak Bhattacharyya\textsuperscript{$\dagger$}}

\address{\textsuperscript{$\diamondsuit$}Samsung Research Institute Bangalore, India\\
\textsuperscript{$\dagger$}Indian Institute of Technology Bombay, India\\
\textsuperscript{$\clubsuit$}IITB-Monash Research Academy, India\\
\textsuperscript{$\star$}Monash University, Australia\\
\textsuperscript{$\heartsuit$}CSIRO Data61, Australia\\
        \textsuperscript{$\diamondsuit$}a.sheoran@samsung.com, \textsuperscript{$\dagger$}\{diptesh, pb\}@cse.iitb.ac.in, \textsuperscript{$\heartsuit$}aditya.joshi@data61.csiro.au\\}

\abstract{
Cross-domain sentiment analysis (CDSA) helps to address the problem of data scarcity in scenarios where labelled data for a domain (known as the target domain) is unavailable or insufficient. However, the decision to choose a domain (known as the source domain) to leverage from is, at best, intuitive. In this paper, we investigate text similarity metrics to facilitate source domain selection for CDSA. We report results on 20 domains (all possible pairs) using 11 similarity metrics. Specifically, we compare CDSA performance with these metrics for different domain-pairs to enable the selection of a suitable source domain, given a target domain. These metrics include two novel metrics for evaluating domain adaptability to help source domain selection of labelled data and utilize word and sentence-based embeddings as metrics for unlabelled data. The goal of our experiments is a recommendation chart that gives the K best source domains for CDSA for a given target domain. We show that the best K source domains returned by our similarity metrics have a precision of over 50\%, for varying values of K. \\ \newline  
}

\begin{document}

\maketitleabstract

\section{Introduction}
Sentiment analysis (SA) deals with automatic detection of opinion orientation in text~\cite{liu2010sentiment}. Domain-specificity of sentiment words, and, as a result, sentiment analysis is also a well-known challenge. A popular example being `\textit{unpredictable}' that is positive for a book review (as in `\textit{The plot of the book is unpredictable}') but negative for an automobile review (as in `\textit{The steering of the car is unpredictable}'). Therefore, a classifier that has been trained on book reviews may not perform as well for automobile reviews~\cite{pang2008opinion}.

However, sufficient datasets may not be available for a domain in which an SA system is to be trained. This has resulted in research in cross-domain sentiment analysis (CDSA). CDSA refers to approaches where the training data is from a different domain (referred to as the `\textit{source domain}') as compared to that of the test data (referred to as the `\textit{target domain}'). \newcite{ben2007analysis} show that similarity between the source and target domains can be used as indicators for domain adaptation, in general.

In this paper, we validate the idea for CDSA. We use similarity metrics as a basis for source domain selection. We implement an LSTM-based sentiment classifier and evaluate its performance for CDSA for a dataset of reviews from twenty domains. We then compare it with similarity metrics to understand which metrics are useful. The resultant deliverable is a recommendation chart of source domains for cross-domain sentiment analysis.\\

The key contributions of this work are:
\begin{enumerate}
    \item We compare eleven similarity metrics (four that use labelled data for the target domain, seven that do not use labelled data for the target domain) with the CDSA performance of 20 domains. Out of these eleven metrics, we introduce two new metrics.  
    \item Based on CDSA results, we create a recommendation chart that prescribes domains that are the best as the source or target domain, for each of the domains. 
    \item In general, we show which similarity metrics are crucial indicators of the benefit to a target domain, in terms of source domain selection for CDSA. 
\end{enumerate}

With rising business applications of sentiment analysis, the convenience of cross-domain adaptation of sentiment classifiers is an attractive proposition. We hope that our recommendation chart will be a useful resource for the rapid development of sentiment classifiers for a domain of which a dataset may not be available. Our approach is based on the hypothesis that if source and target domains are similar, their CDSA accuracy should also be higher given all other conditions (such as data size) are the same. The rest of the paper is organized as follows. We describe related work in Section~\ref{sec:relwork} We then introduce our sentiment classifier in Section \ref{sec:sentclass} and the similarity metrics in Section~\ref{sec:simmet} The results are presented in Section \ref{sec:res} followed by a discussion in Section~\ref{sec:disc} Finally, we conclude the paper in Section~\ref{sec:concl}

\section{Related Work}
\label{sec:relwork}
Cross-domain adaptation has been reported for several NLP tasks such as part-of-speech tagging~\cite{blitzer2006domain}, dependency parsing~\cite{zhang2009cross}, and named entity recognition~\cite{daume2007frustratingly}. Early work in CDSA is by \newcite{denecke2009sentiwordnet}. They show that lexicons such as SentiWordnet do not perform consistently for sentiment classification of multiple domains. Typical statistical approaches for CDSA use active learning~\cite{li2013active}, co-training~\cite{chen2011co} or spectral feature alignment~\cite{pan2010cross}.  In terms of the use of topic models for CDSA, \newcite{he2011automatically} adapt the joint sentiment tying model by introducing domain-specific sentiment-word priors. Similarly, cross-domain sentiment and topic lexicons have been extracted using automatic methods~\cite{li2012cross}.  \newcite{glorot2011domain} present a method for domain adaptation of sentiment classification that uses deep architectures. Our work differs from theirs in terms of computational intensity (deep architecture) and scale (4 domains only).

In this paper, we compare similarity metrics with cross-domain adaptation for the task of sentiment analysis. This has been performed for several other tasks. Recent work by \newcite{dai2019using} uses similarity metrics to select the domain from which pre-trained embeddings should be obtained for named entity recognition. Similarly, \newcite{schultz2018distance} present a method for source domain selection as a weighted sum of similarity metrics. They use statistical classifiers such as logistic regression and support vector machines. However, the similarity measures used are computationally intensive. To the best of our knowledge, this is the first work at this scale that compares different \textit{cost-effective} similarity metrics with the performance of CDSA.
 
\section{Sentiment Classifier}
\label{sec:sentclass}
The core of this work is a sentiment classifier for different domains. We use the DRANZIERA benchmark dataset \cite{dragoni2016dranziera}, which consists of Amazon reviews from 20 domains such as automatives, baby products, beauty products, \textit{etc.} The detailed list can be seen in Table 1. To ensure that the datasets are balanced across all domains, we randomly select 5000 positive and 5000 negative reviews from each domain. The length of the reviews ranges from 5 words to 1654 words across all domains, with an average length ranging from 71 words to 125 words per domain. We point the reader to the original paper for detailed dataset statistics.

We normalize the dataset by removing numerical values, punctuations, stop words, and changing all words to the lower case. To train the sentiment classifier, we use an LSTM-based sentiment classifier. It consists of an embedding layer initialized with pre-trained GloVe word embeddings of 100 dimensions. We specify a hidden layer with 128 units and maintain the batch size at 300\footnote{A smaller batch size results in better accuracy, but at the cost of time. Since the purpose is only to compare the domains and we deal with 400 domain-pairs, we use a larger batch size to save time.}.  We train this model for 20 epochs with a dropout factor of 0.2 and use \textit{sigmoid} as the activation function. For \textit{In-domain sentiment analysis}, we report a 5-fold classification accuracy with a train-test split of 8000 and 2000 reviews. In \textit{cross-domain set up}, we report an average accuracy over 5 splits of 2000 reviews in the target domain in Table \ref{tab:CDSA}.

\begin{table*}[t!]
\centering
\begin{adjustbox}{max width=\textwidth}{}
\begin{tabular}{|c|c|c|c|c|c|c|c|c|c|c|c|c|c|c|c|c|c|c|c|c|}
\hline
                              & \textbf{D1} & \textbf{D2} & \textbf{D3} & \textbf{D4} & \textbf{D5} & \textbf{D6} & \textbf{D7} & \textbf{D8} & \textbf{D9} & \textbf{D10} & \textbf{D11} & \textbf{D12} & \textbf{D13} & \textbf{D14} & \textbf{D15} & \textbf{D16} & \textbf{D17} & \textbf{D18} & \textbf{D19} & \textbf{D20} \\ \hline
\textbf{D1}  & 84.84                        & 70.15                        & 72.58                        & 73.94                        & \textbf{76.92}               & 74.61                        & 70.34                        & 71.06                        & 77.78                        & \textbf{82.83}                & \textbf{76.94}                & 72.93                         & 70.64                         & 69.74                         & 73.56                         & 73.03                         & 74.41                         & 70.19                         & 77.85                         & 76.81                         \\ \hline
\textbf{D2}  & 76.34                        & 83.24                        & 77.71                        & 77.83                        & 71.28                        & \textbf{81.53}               & 80.15                        & 76.61                        & 81.63                        & 75.62                         & 68.03                         & 80.85                         & 79.63                         & 76.45                         & 82.69                         & 77.64                         & 81.49                         & 79.86                         & 80.58                         & 74.62                         \\ \hline
\textbf{D3}  & 74.47                        & 75.00                        & 85.78                        & \textbf{80.16}               & 68.12                        & 77.38                        & 76.06                        & 74.47                        & 80.83                        & 73.03                         & 67.58                         & 78.21                         & 74.35                         & 74.24                         & 75.76                         & 75.59                         & 78.44                         & 74.79                         & 78.97                         & 72.93                         \\ \hline
\textbf{D4}  & 75.66                        & 74.32                        & \textbf{80.31}               & 84.49                        & 68.01                        & 79.01                        & 75.66                        & \textbf{78.60}               & 82.08                        & 75.87                         & 73.14                         & 77.64                         & 77.71                         & 77.08                         & 82.18                         & 75.02                         & 80.73                         & 77.38                         & 80.54                         & 75.05                         \\ \hline
\textbf{D5}  & 81.14                        & 69.81                        & 70.86                        & 73.09                        & 81.56                        & 73.93                        & 65.89                        & 69.26                        & 73.42                        & 79.91                         & 74.97                         & 72.43                         & 71.93                         & 71.67                         & 78.53                         & 74.99                         & 73.38                         & 79.93                         & 77.22                         & 75.20                         \\ \hline
\textbf{D6}  & 74.19                        & 73.87                        & 71.99                        & 76.37                        & 67.03                        & 93.67                        & 73.14                        & 74.80                        & 78.95                        & 75.39                         & 67.55                         & 75.15                         & 75.58                         & 69.90                         & \textbf{83.76}                & 70.99                         & \textbf{82.53}                & 73.88                         & 77.89                         & 70.87                         \\ \hline
\textbf{D7}  & 75.57                        & 77.93                        & 75.67                        & 75.08                        & 67.83                        & 80.60                        & 85.16                        & 73.51                        & 83.47                        & 74.93                         & 72.62                         & 80.65                         & 77.70                         & 71.34                         & 79.64                         & 77.07                         & 79.55                         & 78.31                         & 79.47                         & 72.81                         \\ \hline
\textbf{D8}  & 73.83                        & 73.29                        & 78.39                        & 79.82                        & 68.64                        & 78.40                        & 72.10                        & 83.74                        & 81.03                        & 72.80                         & 68.99                         & 77.73                         & 77.83                         & \textbf{78.52}                & 83.08                         & 74.57                         & 77.54                         & 75.25                         & 80.74                         & 74.55                         \\ \hline
\textbf{D9}  & 75.39                        & 72.90                        & 78.70                        & 76.93                        & 68.48                        & 78.65                        & 79.40                        & 76.42                        & 87.01                        & 74.86                         & 73.07                         & 79.22                         & 78.29                         & 73.17                         & 81.13                         & 76.75                         & 79.94                         & 77.09                         & 81.80                         & 75.46                         \\ \hline
\textbf{D10} & \textbf{82.75}               & 72.69                        & 73.83                        & 73.59                        & 75.46                        & 76.39                        & 71.69                        & 69.29                        & 76.86                        & 84.50                         & 76.59                         & 71.96                         & 71.46                         & 70.18                         & 77.63                         & 73.14                         & 74.45                         & 72.74                         & 77.43                         & 75.78                         \\ \hline
\textbf{D11} & 77.11                        & 65.46                        & 66.81                        & 72.53                        & 72.59                        & 71.69                        & 64.73                        & 69.97                        & 76.28                        & 77.77                         & 84.98                         & 69.19                         & 68.21                         & 65.39                         & 79.05                         & 68.69                         & 71.89                         & 66.95                         & 78.45                         & 76.76                         \\ \hline
\textbf{D12} & 78.31                        & 79.11                        & 78.49                        & 78.69                        & 69.70                        & 80.15                        & 80.10                        & 74.63                        & \textbf{83.58}               & 76.43                         & 72.59                         & 84.79                         & 79.92                         & 76.94                         & 81.45                         & \textbf{78.52}                & 81.04                         & 80.62                         & 80.73                         & 77.5                          \\ \hline
\textbf{D13} & 76.00                        & 79.46                        & 77.00                        & 78.42                        & 70.22                        & 81.12                        & 79.35                        & 75.07                        & 82.86                        & 75.98                         & 70.71                         & 81.23                         & 83.74                         & 76.04                         & 78.92                         & 78.11                         & 81.33                         & \textbf{81.89}                & 81.19                         & 75.28                         \\ \hline
\textbf{D14} & 74.28                        & 77.31                        & 80.29                        & 78.66                        & 68.89                        & 78.32                        & 75.50                        & 77.77                        & 80.99                        & 72.3                          & 69.51                         & 79.19                         & 79.26                         & 84.81                         & 78.85                         & 75.77                         & 78.28                         & 77.91                         & 79.68                         & 73.09                         \\ \hline
\textbf{D15} & 71.91                        & 72.34                        & 71.26                        & 75.29                        & 65.59                        & 77.28                        & 73.09                        & 69.79                        & 77.22                        & 72.34                         & 68.52                         & 73.94                         & 70.79                         & 67.61                         & 95.48                         & 69.78                         & 77.51                         & 71.08                         & 78.00                         & 70.31                         \\ \hline
\textbf{D16} & 72.15                        & 75.18                        & 76.59                        & 75.44                        & 71.09                        & 75.96                        & 77.94                        & 72.14                        & 79.58                        & 72.47                         & 68.65                         & 79.89                         & 77.45                         & 72.78                         & 75.79                         & 84.49                         & 76.62                         & 76.09                         & 77.79                         & 77.07                         \\ \hline
\textbf{D17} & 77.14                        & 77.22                        & 77.27                        & 77.06                        & 69.69                        & 84.02                        & 76.59                        & 75.45                        & 81.09                        & 75.66                         & 72.78                         & 78.66                         & 78.68                         & 71.81                         & 81.93                         & 76.02                         & 88.18                         & 77.87                         & 81.19                         & 73.93                         \\ \hline
\textbf{D18} & 77.15                        & \textbf{80.04}               & 76.21                        & 79.09                        & 72.11                        & 80.09                        & \textbf{80.92}               & 75.52                        & 82.29                        & 75.59                         & 68.54                         & \textbf{81.92}                & \textbf{82.95}                & 76.06                         & 81.38                         & 78.66                         & 81.71                         & 82.18                         & 81.14                         & 75.61                         \\ \hline
\textbf{D19} & 78.83                        & 71.26                        & 75.33                        & 76.18                        & 71.94                        & 77.90                        & 73.14                        & 71.98                        & 82.17                        & 77.20                         & 74.09                         & 75.36                         & 75.99                         & 73.53                         & 82.69                         & 74.35                         & 76.76                         & 75.29                         & 86.77                         & \textbf{78.85}                \\ \hline
\textbf{D20} & 79.08                        & 70.15                        & 71.98                        & 73.72                        & 72.27                        & 77.21                        & 66.81                        & 70.14                        & 79.33                        & 78.92                         & 74.29                     & 69.83                         & 74.32                         & 70.09                         & 80.37                         & 69.79                         & 74.83                         & 70.38                         & \textbf{81.19}                & 83.19                         \\ \hline
\end{tabular}%
\end{adjustbox}
\caption{Accuracy percentage for all train-test pairs. Domains on rows are source domains and columns are target domains. Domain labels are D1: Amazon Instant Video, D2: Automotive, D3: Baby, D4: Beauty, D5: Books, D6: Clothing Accessories, D7: Electronics, D8: Health, D9: Home, D10:  Kitchen, D11: Movies TV, D12: Music, D13: Office Products, D14: Patio, D15: Pet Supplies, D15: Shoes, D16: Software, D17: Sports Outdoors, D18: Tools Home Improvement, D19: Toys Games, D20: Video Games.}
\label{tab:CDSA}
\end{table*}

\begin{table*}[]
\centering
\resizebox{0.99\textwidth}{!}{%
\begin{tabular}{|c|c|c|c|c|c|c|c|c|c|c|c|c|c|c|c|c|c|c|c|c|}
\hline
 & D1 & D2 & D3 & D4 & D5 & D6 & D7 & D8 & D9 & D10 & D11 & D12 & D13 & D14 & D15 & D16 & D17 & D18 & D19 & D20 \\ \hline
D1 & - & 0.27 & 0.27 & 0.40 & 0.23 & 0.50 & 0.37 & 0.20 & 0.57 & 0.20 & 0.33 & 0.20 & 0.30 & 0.30 & 0.20 & 0.07 & 0.23 & 0.20 & 0.20 & 0.87 \\ \hline
D2 & - & - & 0.53 & 0.37 & 0.27 & 0.23 & 0.43 & 0.30 & 0.30 & 0.30 & 0.40 & 0.33 & 0.43 & 0.40 & 0.40 & 0.20 & 0.33 & 0.50 & 0.43 & 0.27 \\ \hline
D3 & - & - & - & 0.30 & 0.30 & 0.27 & 0.63 & 0.33 & 0.30 & 0.33 & 0.67 & 0.23 & 0.50 & 0.50 & 0.63 & 0.30 & 0.47 & 0.53 & 0.53 & 0.27 \\ \hline
D4 & - & - & - & - & 0.27 & 0.37 & 0.30 & 0.17 & 0.37 & 0.20 & 0.30 & 0.17 & 0.30 & 0.33 & 0.23 & 0.07 & 0.27 & 0.27 & 0.27 & 0.40 \\ \hline
D5 & - & - & - & - & - & 0.20 & 0.33 & 0.23 & 0.23 & 0.23 & 0.37 & 0.17 & 0.37 & 0.43 & 0.37 & 0.13 & 0.43 & 0.30 & 0.40 & 0.23 \\ \hline
D6 & - & - & - & - & - & - & 0.33 & 0.17 & 0.37 & 0.20 & 0.30 & 0.17 & 0.27 & 0.27 & 0.20 & 0.07 & 0.23 & 0.20 & 0.20 & 0.47 \\ \hline
D7 & - & - & - & - & - & - & - & 0.33 & 0.37 & 0.37 & 0.73 & 0.33 & 0.73 & 0.57 & 0.63 & 0.30 & 0.53 & 0.50 & 0.60 & 0.33 \\ \hline
D8 & - & - & - & - & - & - & - & - & 0.27 & 0.30 & 0.37 & 0.30 & 0.33 & 0.47 & 0.33 & 0.20 & 0.27 & 0.30 & 0.33 & 0.20 \\ \hline
D9 & - & - & - & - & - & - & - & - & - & 0.20 & 0.37 & 0.17 & 0.33 & 0.30 & 0.23 & 0.13 & 0.23 & 0.27 & 0.27 & 0.60 \\ \hline
D10 & - & - & - & - & - & - & - & - & - & - & 0.37 & 0.43 & 0.40 & 0.33 & 0.33 & 0.17 & 0.37 & 0.37 & 0.47 & 0.20 \\ \hline
D11 & - & - & - & - & - & - & - & - & - & - & - & 0.30 & 0.73 & 0.57 & 0.60 & 0.30 & 0.50 & 0.50 & 0.53 & 0.33 \\ \hline
D12 & - & - & - & - & - & - & - & - & - & - & - & - & 0.37 & 0.27 & 0.27 & 0.10 & 0.30 & 0.30 & 0.40 & 0.20 \\ \hline
D13 & - & - & - & - & - & - & - & - & - & - & - & - & - & 0.57 & 0.50 & 0.30 & 0.47 & 0.53 & 0.53 & 0.30 \\ \hline
D14 & - & - & - & - & - & - & - & - & - & - & - & - & - & - & 0.57 & 0.23 & 0.43 & 0.43 & 0.53 & 0.30 \\ \hline
D15 & - & - & - & - & - & - & - & - & - & - & - & - & - & - & - & 0.30 & 0.47 & 0.43 & 0.60 & 0.23 \\ \hline
D16 & - & - & - & - & - & - & - & - & - & - & - & - & - & - & - & - & 0.27 & 0.23 & 0.27 & 0.07 \\ \hline
D17 & - & - & - & - & - & - & - & - & - & - & - & - & - & - & - & - & - & 0.40 & 0.53 & 0.27 \\ \hline
D18 & - & - & - & - & - & - & - & - & - & - & - & - & - & - & - & - & - & - & 0.50 & 0.23 \\ \hline
D19 & - & - & - & - & - & - & - & - & - & - & - & - & - & - & - & - & - & - & - & 0.23 \\ \hline
\end{tabular}%
}
\caption{N-grams co-occurrence matrix depicting the percent point match among the top-10 bigrams, trigrams and quadgrams for the data used in each domain. Domain labels are D1: Amazon Instant Video, D2: Automotive, D3: Baby, D4: Beauty, D5: Books, D6: Clothing Accessories, D7: Electronics, D8: Health, D9: Home, D10:  Kitchen, D11: Movies TV, D12: Music, D13: Office Products, D14: Patio, D15: Pet Supplies, D15: Shoes, D16: Software, D17: Sports Outdoors, D18: Tools Home Improvement, D19: Toys Games, D20: Video Games.}
\label{tab:n-grams}
\end{table*}

\section{Similarity Metrics}
\label{sec:simmet}
In table \ref{tab:n-grams}, we present the n-gram percent match among the domain data used in our experiments. We observe that the n-gram match from among this corpora is relatively low and simple corpus similarity measures which use orthographic techniques cannot be used to obtain domain similarity. Hence, we propose the use of the metrics detailed below to perform our experiments.

We use a total of 11 metrics over two scenarios: the first that uses labelled data, while the second that uses unlabelled data.
\begin{enumerate}
    \item \textbf{Labelled Data:} Here, each review in the target domain data is labelled either positive or negative, and a number of such labelled reviews are insufficient in size for training an efficient model.
    \item \textbf{Unlabelled Data:} Here, positive and negative labels are absent from the target domain data, and the number of such reviews may or may not be sufficient in number.
\end{enumerate}

\noindent We explain all our metrics in detail later in this section. These 11 metrics can also be classified into two categories: 
\begin{itemize}
    \item \textbf{Symmetric Metrics} - The metrics which consider domain-pairs $(D_1,D_2)$ and $(D_2,D_1)$ as the same and provide similar results for them \textit{viz.} Significant Words Overlap, Chameleon Words Similarity, Symmetric KL Divergence, Word2Vec embeddings, GloVe embeddings, FastText word embeddings, ELMo based embeddings and Universal Sentence Encoder based embeddings. 
    \item \textbf{Asymmetric Metrics} - The metrics which are 2-way in nature \textit{i.e.,} $(D_1,D_2)$ and $(D_2,D_1)$ have different similarity values \textit{viz.} Entropy Change, Doc2Vec embeddings, and FastText sentence embeddings. These metrics offer additional advantage as they can help decide which domain to train from and which domain to test on amongst $D_1$ and $D_2$.
\end{itemize}
\subsection{Metrics: Labelled Data}

Training models for prediction of sentiment can cost one both valuable time and resources. The availability of pre-trained models is \textit{cost-effective} in terms of both time and resources. One can always train new models and test for each source domain since labels are present for the source domain data. However, it is feasible only when trained classification models are available for all source domains. If pre-trained models are unavailable, training for each source domain can be highly intensive both in terms of time and resources. This makes it important to devise \textit{easy-to-compute} metrics that use labelled data in the source and target domains.  

When target domain data is labelled, we use the following four metrics for comparing and ranking source domains for a particular target domain:

\subsubsection*{LM1: Significant Words Overlap}

All words in a domain are not significant for sentiment expression. For example, \textit{comfortable} is significant in the `Clothing' domain but not as significant in the `Movie' domain. In this metric, we build upon existing work by \newcite{sharma2018identifying} and extract significant words from each domain using the $\chi^2$ test. This method relies on computing the statistical significance of a word based on the polarity of that word in the domain. For our experiments, we consider only the words which appear at least 10 times in the corpus and have a $\chi^2$ value greater than or equal to 1. The $\chi^2$ value is calculated as follows:

\begin{equation}
    \chi^2(w) = \frac{({c_p}^w - \mu^w)^2 + ({c_n}^w - \mu^w)^2}{\mu^w}
\end{equation}

Where ${c_p}^w$ and ${c_n}^w$ are the observed counts of word $w$ in positive and negative reviews, respectively.  $\mu^w$ is the expected count, which is kept as half of the total number of occurrences of $w$ in the corpus.
We hypothesize that, if a domain-pair $(D_1,D_2)$ shares a larger number of significant words than the pair $(D_1,D_3)$, then $D_1$ is closer to $D_2$ as compared to $D_3$, since they use relatively higher number of similar words for sentiment expression. For every target domain, we compute the intersection of significant words with all other domains and rank them on the basis of intersection count. The utility of this metric is that it can also be used in a scenario where target domain data is unlabelled, but source domain data is labelled. It is due to the fact that once we obtain significant words in the source domain, we just need to search for them in the target domain to find out common significant words.

\subsubsection*{LM2: Symmetric KL-Divergence (SKLD)}

KL Divergence can be used to compare the probabilistic distribution of polar words in two domains~\cite{kullback1951information}. A lower KL Divergence score indicates that the probabilistic distribution of polar words in two domains is identical. This implies that the domains are close to each other, in terms of sentiment similarity. Therefore, to rank source domains for a target domain using this metric, we inherit the concept of symmetric KL Divergence proposed by \newcite{murthy2018judicious} and use it to compute average Symmetric KL-Divergence of common polar words shared by a domain-pair. We label a word as `polar' for a domain if,
\begin{equation}
     |P-N|>=0.5
\end{equation}
where $P$ is the probability of a word appearing in a review which is labelled positive and $N$ is the probability of a word appearing in a review which is labelled negative. 

\noindent SKLD of a polar word for domain-pair $(D_1,D_2)$ is calculated as: 
\begin{equation}
 A \ = \  N_1*log\Bigg(\frac{N_1}{N_2}\Bigg) \ + \ P_1*log\Bigg(\frac{P_1}{P_2}\Bigg)
\end{equation}
\begin{equation}
     B \ = \  N_2*log\Bigg(\frac{N_2}{N_1}\Bigg) \ + \ P_2*log\Bigg(\frac{P_2}{P_1}\Bigg)
\end{equation}
\begin{equation}
    SKLD = \frac{A+B}{2}
\end{equation}

\noindent where $P_i$ and $N_i$ are probabilities of a word appearing under positively labelled and negatively labelled reviews, respectively, in domain $i$. We then take an average of all common polar words.

We observe that, on its own, this metric performs rather poorly. Upon careful analysis of results, we concluded that the imbalance in the number of polar words being shared across domain-pairs is a reason for poor performance. To mitigate this, we compute a confidence term for a domain-pair $(D_1,D_2)$ using the \textit{Jaccard Similarity Coefficient} which is calculated as follows: 
\begin{equation}
    J=\frac{C}{W_1 + W_2 - C}
\end{equation}
where $C$ is the number of common polar words and $W_1$ and $W_2$ are number of polar words in $D_1$ and $D_2$ respectively. The intuition behind this being that the domain-pairs having higher percentage of polar words overlap should be ranked higher compared to those having relatively higher number of polar words. For example, we prefer $(C:40,W_1 :50,W_2 :50)$ over $(C:200,W_1 :500,W_2 :500)$ even though $200$ is greater than $40$.
To compute the final similarity value, we add the reciprocal of $J$ to the SKLD value since a larger value of $J$ will add a smaller fraction to SLKD value. For a smaller SKLD value, the domains would be relatively more similar. This is computed as follows:
\begin{equation}
    (SKLD)_{avg} + \frac{1}{J}
\end{equation}

Domain pairs are ranked in increasing order of this similarity value. After the introduction of the confidence term, a significant improvement in the results is observed.  

\subsubsection*{LM3: Chameleon Words Similarity}

This metric is our novel contribution for domain adaptability evaluation. It helps in detection of `\textit{Chameleon Word}(s)' which change their polarity across domains \cite{sharma2013detecting}. The motivation comes from the fact that chameleon words directly affect the CDSA accuracy. For example, \textit{poignant} is positive in movie domain whereas negative in many other domains \textit{viz.} Beauty, Clothing \textit{etc.} \\
For every common polar word between two domains, $L_1 \ Distance$ between two vectors $[P_1,N_1]$ and $[P_2,N_2]$ is calculated as;
\begin{equation}
    |P_1-P_2|+|N_1-N_2|
\end{equation}
The overall distance is an average overall common polar words. Similar to SKLD, the confidence term based on \textit{Jaccard Similarity Coefficient} is used to counter the imbalance of common polar word count between domain-pairs.  
\begin{equation}
    (L_1 \ Distance)_{avg} + \frac{1}{J}
\end{equation}

\noindent Domain pairs are ranked in increasing order of final value. 

\subsubsection*{LM4: Entropy Change}

Entropy is the degree of randomness. A relatively lower change in entropy, when two domains are concatenated, indicates that the two domains contain similar topics and are therefore closer to each other. This metric is also our novel contribution. Using this metric, we calculate the percentage change in the entropy when the target domain is concatenated with the source domain. We calculate the entropy as the combination of entropy for unigrams, bigrams, trigrams, and quadrigrams. We consider only polar words for unigrams. For bi, tri and quadrigrams, we give priority to polar words by using a \textit{weighted entropy function} and this weighted entropy $E$ is calculated as:
\begin{equation}
    P = -\sum_{i \ \epsilon \ \{X\}}p(X_i)logp(X_i)*w 
\end{equation}
\noindent
\begin{equation}
   Q = -\sum_{j \ \epsilon \ \{Y\}}p(Y_j)logp(Y_j)*\frac{1}{w}
\end{equation}
\begin{equation}
   E = P + Q
\end{equation}

\noindent Here, $X$ is the set of n-grams that contain at least one polar word, $Y$ is the set of n-grams which do not contain any polar word, and $w$ is the weight. For our experiments, we keep the value of $w$ as 1 for unigrams and 5\footnote{We observe that any value of $w$ does not change the relative ranking of domains.} for bi, tri, and quadrigrams. 

We then say that a source domain $D_2$ is more suitable for target domain $D_1$ as compared to source domain $D_3$ if;
\begin{align}
    \Delta E(D_2,D_2+D_1) < \Delta E(D_3,D_3+D_1)
\end{align}
where $D_2+D_1$ indicates combined data obtained by mixing $D_1$ in $D_2$ and $\Delta E$ indicates percentage change in entropy before and after mixing of source and target domains.

Note that this metric offers the advantage of asymmetricity, unlike the other three metrics for labelled data.

\subsection{Metrics: Unlabelled Data}
For unlabelled target domain data, we utilize word and sentence embeddings-based similarity as a metric and use various embedding models. To train word embedding based models, we use Word2Vec \cite{mikolov2013efficient}, GloVe \cite{pennington2014glove}, FastText \cite{bojanowski2017enriching}, and ELMo \cite{peters2018deep}. We also exploit sentence vectors from models trained using Doc2Vec \cite{le2014distributed}, FastText, and Universal Sentence Encoder \cite{cer2018universal}. In addition to using plain sentence vectors, we account for sentiment in sentences using SentiWordnet \cite{baccianella2010sentiwordnet}, where each review is given a sentiment score by taking harmonic mean over scores (obtained from SentiWordnet) of words in a review\footnote{\href{https://github.com/anelachan/sentimentanalysis/blob/master/sentiment.py}{Github: Sentiment Classifier}}.

\subsubsection*{ULM1: Word2Vec}
We train SKIPGRAM models on all the domains to obtain word embeddings. We build models with 50 dimensions\footnote{We train the models with different dimensions and compute scores for dimensions 50, 100, 200, and 300. We choose the dimension which gives us the best results and report it in the following sections, for each metric below.} where the context window is chosen to be 5. For each domain pair, we then compare embeddings of common adjectives in both the domains by calculating \textit{Angular Similarity} \cite{cer2018universal}. It was observed that cosine similarity values were very close to each other, making it difficult to clearly separate domains. Since Angular Similarity distinguishes nearly parallel vectors much better, we use it instead of \textit{Cosine Similarity}. We obtain a similarity value by averaging over all common adjectives. For the final similarity value of this metric, we use \textit{Jaccard Similarity Coefficient} here as well:
\begin{equation}
\label{eq:word2vec}
    (Angular \ Similarity)_{avg} + J
\end{equation}

For a target domain, source domains are ranked in decreasing order of final similarity value.

\begin{figure}[ht!]
  \includegraphics[width=\columnwidth,height=6cm]{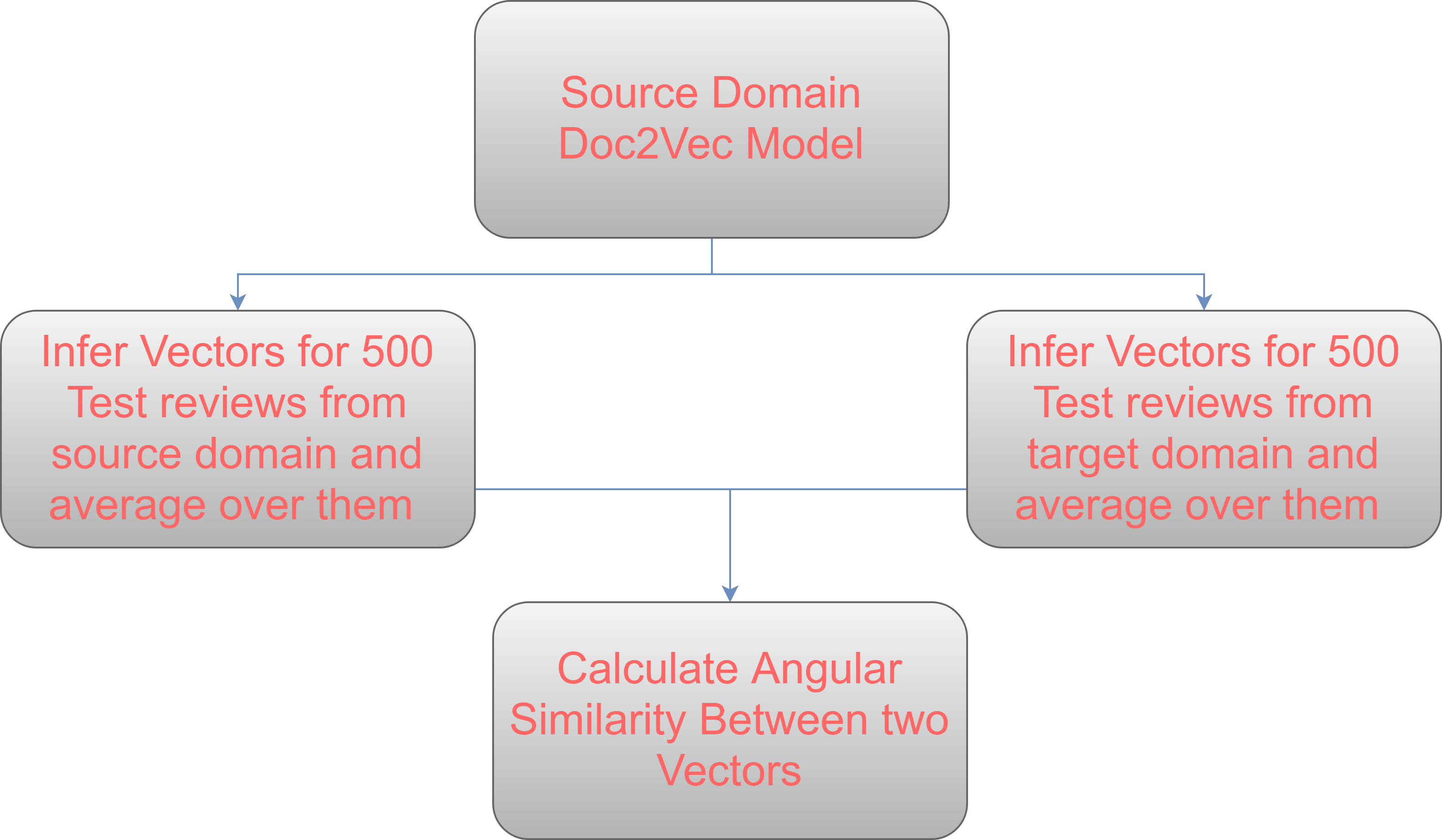}
  \caption{Experimental Setup for Doc2Vec}
\label{fig:Doc2Vec}
\end{figure}

\begin{table*}[t!]
\centering
\resizebox{0.85\textwidth}{!}{
\begin{tabular}{|c|c|c|c|c|c|}
\hline
\multicolumn{2}{|c|}{\textbf{Domains}} & \textbf{\begin{tabular}[c]{@{}c@{}}In Domain \\ Accuracy (\%)\end{tabular}} & \textbf{\begin{tabular}[c]{@{}c@{}}Average CDSA \\ Accuracy Degradation (\%)\end{tabular}} & \textbf{\begin{tabular}[c]{@{}c@{}}Best Source \\ Domain\end{tabular}} & \textbf{\begin{tabular}[c]{@{}c@{}}Best Target \\ Domain\end{tabular}} \\ \hline
\textbf{D1} & \textbf{Amazon Instant Video} & 84.84 & 10.82 & D10 & D10 \\ \hline
\textbf{D2} & \textbf{Automotive} & 83.24 & 5.32 & D18 & D15 \\ \hline
\textbf{D3} & \textbf{Baby} & 85.78 & 10.50 & D4 & D9 \\ \hline
\textbf{D4} & \textbf{Beauty} & 84.49 & 7.33 & D3 & D9 \\ \hline
\textbf{D5} & \textbf{Books} & 81.56 & 7.48 & D1 & D1 \\ \hline
\textbf{D6} & \textbf{Clothing Accessories} & 93.67 & 19.26 & D2 & D15 \\ \hline
\textbf{D7} & \textbf{Electronics} & 85.16 & 8.65 & D18 & D9 \\ \hline
\textbf{D8} & \textbf{Health} & 83.74 & 7.58 & D4 & D15 \\ \hline
\textbf{D9} & \textbf{Home Kitchen} & 87.01 & 10.29 & D12 & D19 \\ \hline
\textbf{D10} & \textbf{Movies TV} & 84.50 & 10.08 & D1 & D1 \\ \hline
\textbf{D11} & \textbf{Music} & 84.98 & 13.43 & D1 & D15 \\ \hline
\textbf{D12} & \textbf{Office Products} & 84.79 & 6.45 & D18 & D9 \\ \hline
\textbf{D13} & \textbf{Patio} & 83.74 & 5.84 & D18 & D18 \\ \hline
\textbf{D14} & \textbf{Pet Supplies} & 84.81 & 8.19 & D8 & D9 \\ \hline
\textbf{D15} & \textbf{Shoes} & 95.48 & 23.18 & D6 & D17 \\ \hline
\textbf{D16} & \textbf{Software} & 84.49 & 9.19 & D12 & D12 \\ \hline
\textbf{D17} & \textbf{Sports Outdoors} & 88.18 & 11.12 & D6 & D6 \\ \hline
\textbf{D18} & \textbf{Tools Home Improvement} & 82.18 & 4.01 & D13 & D13 \\ \hline
\textbf{D19} & \textbf{Toys Games} & 86.77 & 10.83 & D20 & D15 \\ \hline
\textbf{D20} & \textbf{Video Games} & 83.19 & 9.26 & D19 & D19 \\ \hline
\end{tabular}%
}
\caption{Our reccomendation chart based on CDSA results: In Domain Accuracy(when source and target domains are same), Average CDSA Accuracy Degradation(average cross-domain testing accuracy loss over all target domains), Best Source Domain, Best Target Domain for each domain.}
\label{cross-domain}
\end{table*}

\subsubsection*{ULM2: Doc2Vec}
Doc2Vec represents each sentence by a dense vector which is trained to predict words in the sentence, given the model. 
It tries to overcome the weaknesses of the bag-of-words model. 
Similar to Word2Vec, we train Doc2Vec models on each domain to extract sentence vectors. We train the models over 100 epochs for 100 dimensions, where the learning rate is 0.025. Since we can no longer leverage adjectives for sentiment, we use SentiWordnet for assigning sentiment scores (ranging from -1 to +1 where -1 denotes a negative sentiment, and +1 denotes a positive sentiment) to reviews (as detailed above) and select reviews which have a score above a certain threshold. We have empirically arrived at $\pm 0.01$ as the threshold value. Any review with a score outside this window is selected. We also restrict the length of reviews to a maximum of 100 words to reduce sparsity.\\\\
After filtering out reviews with sentiment score less than the threshold value, we are left with a minimum of 8000 reviews per domain. We train on 7500 reviews form each domain and test on 500 reviews. To compare a domain-pair $(D_1,D_2)$ where $D_1$ is the source domain and $D_2$ is the target domain, we compute Angular Similarity between two vectors $V_1$ and $V_2$. $V_1$ is obtained by taking an average over 500 test vectors (from $D_1$) inferred from the model trained on $D_1$. $V_2$ is obtained in a similar manner, except that the test data is from $D_2$. Figure \ref{fig:Doc2Vec} shows the experimental setup for this metric.

\subsubsection*{ULM3: GloVe}
Both Word2Vec and GloVe learn vector representations of words from their co-occurrence information. However, GloVe is different in the sense that it is a \textit{count-based} model. 
In this metric, we use GloVe embeddings for adjectives shared by domain-pairs. We train GloVe models for each domain over 50 epochs, for 50 dimensions with a learning rate of 0.05. For computing similarity of a domain-pair, we follow the same procedure as described under the Word2Vec metric. The final similarity value is obtained using equation (\ref{eq:word2vec}).

\subsubsection*{ULM4 and ULM5: FastText}

We train monolingual word embeddings-based models for each domain using the FastText library\footnote{\href{https://github.com/facebookresearch/fastText}{Github: FastText}}. We train these models with 100 dimensions and 0.1 as the learning rate. The size of the context window is limited to 5 since FastText also uses sub-word information. Our model takes into account character n-grams from 3 to 6 characters, and we train our model over 5 epochs. We use the default loss function (\textit{softmax}) for training.

We devise \textbf{two different metrics} out of FastText models to calculate the similarity between domain-pairs. In the first metric (\textbf{ULM4}), we compute the Angular Similarity between the word vectors for all the common adjectives, and for each domain pair just like Word2Vec and GloVe. Overall, similarity for a domain pair is calculated using equation (\ref{eq:word2vec}). As an additional metric (\textbf{ULM5}), we extract sentence vectors for reviews and follow a procedure similar to Doc2Vec. SentiWordnet is used to filter out train and test data using the same threshold window of $\pm 0.01$.

\begin{table*}[ht!]
\centering
\begin{adjustbox}{max width=\textwidth}{}
\begin{tabular}{ccccccccc}
\hline
\multicolumn{1}{|c|}{}     & \multicolumn{2}{c|}{\textbf{Top 3}}                                                                                          & \multicolumn{2}{c|}{\textbf{Top 5}}                                                                                          & \multicolumn{2}{c|}{\textbf{Top 7}}                                                                                          & \multicolumn{2}{c|}{\textbf{Top 10}}                                                                                         \\ \hline
\multicolumn{1}{|c|}{}     & \multicolumn{1}{c|}{\textbf{\begin{tabular}[c]{@{}c@{}}Precision (\%)\end{tabular}}} & \multicolumn{1}{c|}{\textbf{NRA}}   & \multicolumn{1}{c|}{\textbf{\begin{tabular}[c]{@{}c@{}}Precision (\%)\end{tabular}}} & \multicolumn{1}{c|}{\textbf{NRA}}   & \multicolumn{1}{c|}{\textbf{\begin{tabular}[c]{@{}c@{}}Precision (\%)\end{tabular}}} & \multicolumn{1}{c|}{\textbf{NRA}}   & \multicolumn{1}{c|}{\textbf{\begin{tabular}[c]{@{}c@{}}Precision (\%)\end{tabular}}} & \multicolumn{1}{c|}{\textbf{NRA}}   \\ \hline
\multicolumn{9}{c}{\textbf{Labelled}}                                                                \\ \hline
\multicolumn{1}{|c|}{LM1}  & \multicolumn{1}{c|}{45.00}                                                             & \multicolumn{1}{c|}{0.200}          & \multicolumn{1}{c|}{54.00}                                                             & \multicolumn{1}{c|}{\textbf{0.190}} & \multicolumn{1}{c|}{64.29}                                                             & \multicolumn{1}{c|}{\textbf{0.150}} & \multicolumn{1}{c|}{72.00}                                                             & \multicolumn{1}{c|}{0.060}          \\ \hline
\multicolumn{1}{|c|}{LM2}  & \multicolumn{1}{c|}{\textbf{46.67}}                                                    & \multicolumn{1}{c|}{0.183}          & \multicolumn{1}{c|}{\textbf{62.00}}                                                    & \multicolumn{1}{c|}{0.150}          & \multicolumn{1}{c|}{62.86}                                                             & \multicolumn{1}{c|}{0.114}          & \multicolumn{1}{c|}{77.50}                                                             & \multicolumn{1}{c|}{0.055}          \\ \hline
\multicolumn{1}{|c|}{LM3}  & \multicolumn{1}{c|}{\textbf{46.67}}                                                    & \multicolumn{1}{c|}{0.183}          & \multicolumn{1}{c|}{\textbf{62.00}}                                                    & \multicolumn{1}{c|}{0.150}          & \multicolumn{1}{c|}{63.57}                                                             & \multicolumn{1}{c|}{0.114}          & \multicolumn{1}{c|}{\textbf{78.00}}                                                    & \multicolumn{1}{c|}{0.055}          \\ \hline
\multicolumn{1}{|c|}{LM4}  & \multicolumn{1}{c|}{\textbf{46.67}}                                                    & \multicolumn{1}{c|}{\textbf{0.233}} & \multicolumn{1}{c|}{59.00}                                                             & \multicolumn{1}{c|}{0.180}          & \multicolumn{1}{c|}{\textbf{69.29}}                                                    & \multicolumn{1}{c|}{0.129}          & \multicolumn{1}{c|}{76.00}                                                             & \multicolumn{1}{c|}{\textbf{0.110}} \\ \hline
\multicolumn{9}{c}{\textbf{Unlabelled}}                                                                                                                                                                                                                                                                                                                                                                                                                                                                                                                \\ \hline
\multicolumn{1}{|c|}{ULM1} & \multicolumn{1}{c|}{51.67}                                                             & \multicolumn{1}{c|}{0.267}          & \multicolumn{1}{c|}{62.00}                                                             & \multicolumn{1}{c|}{\textbf{0.210}} & \multicolumn{1}{c|}{66.43}                                                             & \multicolumn{1}{c|}{\textbf{0.186}} & \multicolumn{1}{c|}{77.50}                                                             & \multicolumn{1}{c|}{0.080}          \\ \hline
\multicolumn{1}{|c|}{ULM2} & \multicolumn{1}{c|}{40.00}                                                             & \multicolumn{1}{c|}{0.100}          & \multicolumn{1}{c|}{49.00}                                                             & \multicolumn{1}{c|}{0.110}          & \multicolumn{1}{c|}{61.43}                                                             & \multicolumn{1}{c|}{0.122}          & \multicolumn{1}{c|}{76.00}                                                             & \multicolumn{1}{c|}{0.100}          \\ \hline
\multicolumn{1}{|c|}{ULM3} & \multicolumn{1}{c|}{41.67}                                                             & \multicolumn{1}{c|}{0.150}          & \multicolumn{1}{c|}{50.00}                                                             & \multicolumn{1}{c|}{0.140}          & \multicolumn{1}{c|}{59.29}                                                             & \multicolumn{1}{c|}{0.136}          & \multicolumn{1}{c|}{73.50}                                                             & \multicolumn{1}{c|}{0.045}          \\ \hline
\multicolumn{1}{|c|}{ULM4} & \multicolumn{1}{c|}{51.67}                                                             & \multicolumn{1}{c|}{0.217}          & \multicolumn{1}{c|}{55.00}                                                             & \multicolumn{1}{c|}{0.170}          & \multicolumn{1}{c|}{62.14}                                                             & \multicolumn{1}{c|}{0.157}          & \multicolumn{1}{c|}{75.50}                                                             & \multicolumn{1}{c|}{0.065}          \\ \hline
\multicolumn{1}{|c|}{ULM5} & \multicolumn{1}{c|}{45.00}                                                             & \multicolumn{1}{c|}{0.267}          & \multicolumn{1}{c|}{54.00}                                                             & \multicolumn{1}{c|}{0.180}          & \multicolumn{1}{c|}{60.71}                                                             & \multicolumn{1}{c|}{0.157}          & \multicolumn{1}{c|}{68.00}                                                             & \multicolumn{1}{c|}{\textbf{0.125}} \\ \hline
\multicolumn{1}{|c|}{ULM6} & \multicolumn{1}{c|}{56.67}                                                             & \multicolumn{1}{c|}{0.233}          & \multicolumn{1}{c|}{63.00}                                                             & \multicolumn{1}{c|}{0.200}          & \multicolumn{1}{c|}{70.00}                                                             & \multicolumn{1}{c|}{0.179}          & \multicolumn{1}{c|}{\textbf{81.00}}                                                    & \multicolumn{1}{c|}{0.070}          \\ \hline
\multicolumn{1}{|c|}{ULM7} & \multicolumn{1}{c|}{\textbf{58.33}}                                                    & \multicolumn{1}{c|}{\textbf{0.300}} & \multicolumn{1}{c|}{\textbf{64.00}}                                                    & \multicolumn{1}{c|}{\textbf{0.210}} & \multicolumn{1}{c|}{\textbf{70.72}}                                                    & \multicolumn{1}{c|}{0.179}          & \multicolumn{1}{c|}{80.00}                                                             & \multicolumn{1}{c|}{0.090}          \\ \hline
\end{tabular}%
\end{adjustbox}
  \caption{Precision and Normalised Ranking Accuracy (NRA) for top-K source domain matching over all domains.}
  \label{tab:results}
\end{table*}

\subsubsection*{ULM6: ELMo}

We use the pre-trained deep contextualized word representation model provided by the ELMo library\footnote{\href{https://github.com/allenai/allennlp}{GitHub: ELMo}}. 
Unlike Word2Vec, GloVe, and FastText,  ELMo gives multiple embeddings for a word based on different contexts it appears in the corpus. 

In ELMo, higher-level LSTM states capture the context-dependent aspects of word meaning. Therefore, we use only the topmost layer for word embeddings with 1024 dimensions. Multiple contextual embeddings of a word are averaged to obtain a single vector. We again use average Angular Similarity of word embeddings for common adjectives to compare domain-pairs along with \textit{Jaccard Similarity Coefficient}. The final similarity value is obtained using equation (\ref{eq:word2vec}).

\subsubsection*{ULM7: Universal Sentence Encoder}

One of the most recent contributions to the area of sentence embeddings is the Universal Sentence Encoder. Its transformer-based sentence encoding model constructs sentence embeddings using the encoding sub-graph of the transformer architecture \cite{vaswani2017attention}. We leverage these embeddings and devise a metric for our work.

We extract sentence vectors of reviews in each domain using tensorflow-hub model toolkit\footnote{\href{https://tfhub.dev/google/universal-sentence-encoder/2}{TensorFlow Hub}}. The dimensions of each vector are 512. To find out the similarity between a domain-pair, we extract top 500 reviews from both domains based on the sentiment score acquired using SentiWordnet (as detailed above) and average over them to get two vectors with 512 dimensions each. After that, we find out the Angular Similarity between these vectors to rank all source domains for a particular target domain in decreasing order of similarity.

\begin{figure*}[ht]
  \includegraphics[width=\textwidth,height=8.5cm]{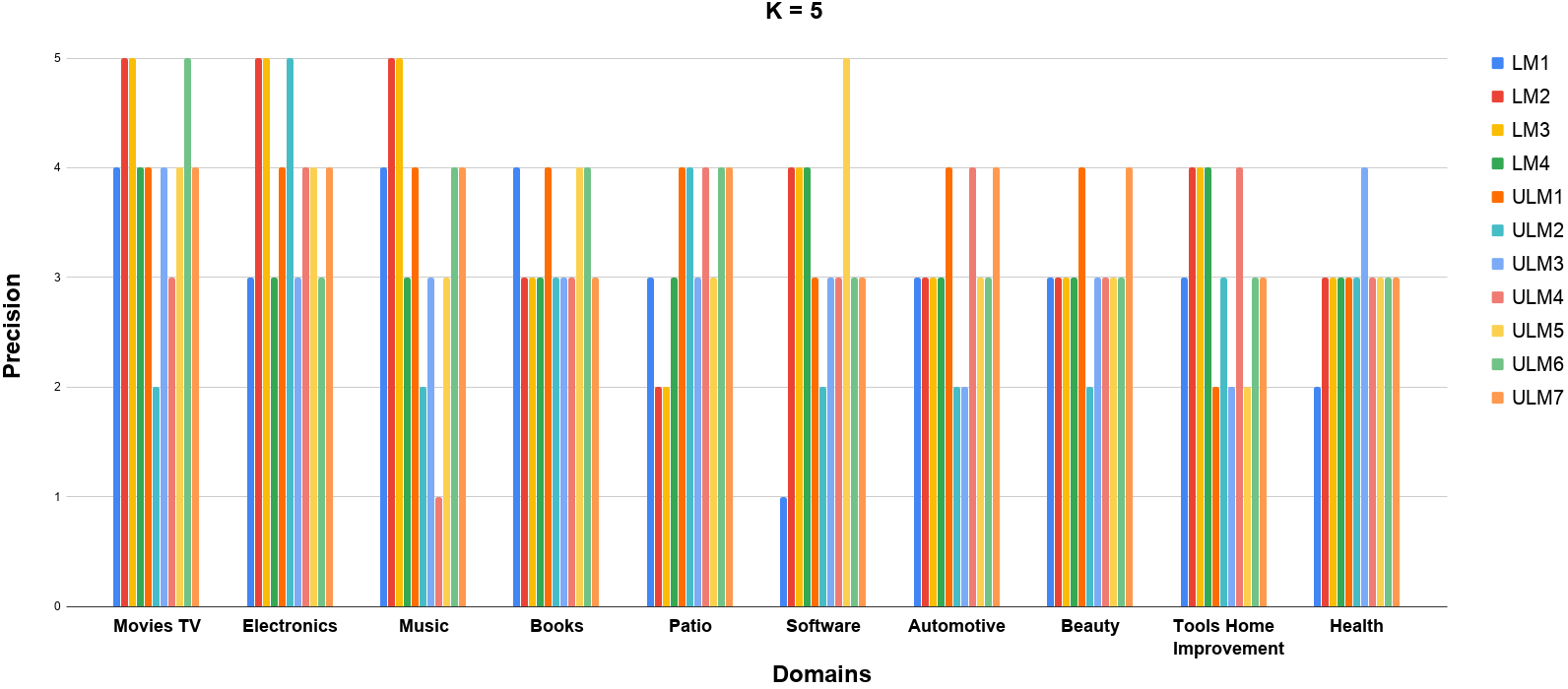}
  \caption{Precision for K=5 over top 10 domains. Precision is the intersection between the top-K source domains predicted by the metric and top-K source domains as per CDSA accuracy.}
\label{fig:overlap}
\end{figure*}

\begin{figure*}[ht!]
  \includegraphics[width=\textwidth,height=8.5cm]{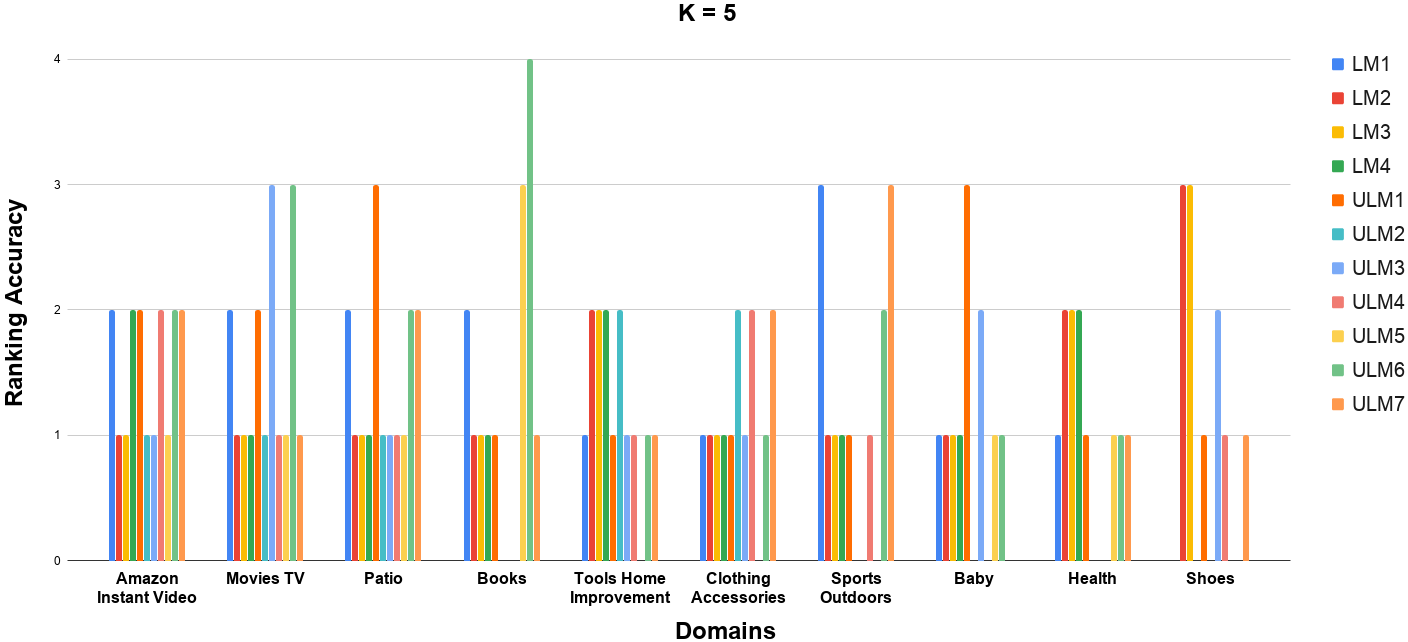}
  \caption{Ranking Accuracy for K=5 over top 10 domains. Ranking accuracy is the number of predicted source domains which are \textbf{ranked correctly} by the metric.}
\label{fig:precision}
\end{figure*}

\section{Results}
\label{sec:res}

We show the results of the classifier's CDSA performance followed by metrics evaluation on the top 10 domains. Finally, we present an overall comparison of metrics for all the domains.

Table \ref{cross-domain} shows the average CDSA accuracy degradation in each domain when it is selected as the source domain, and the rest of the domains are selected as the target domain. We also show in-domain sentiment analysis accuracy, the best source domain (on which CDSA classifier is trained), and the best target domain (on which CDSA classifier is tested) in the table. D15 suffers from the maximum average accuracy degradation, and D18 performs the best with least average accuracy degradation, which is also supported by its number of appearances \textit{i.e.,} 4, as the best source domain in the table. As for the best target domain, D9 appears the maximum number of times.

To compare metrics, we use two parameters: \textit{Precision} and \textit{Ranking Accuracy}.
\begin{itemize}
    \item \textbf{Precision:} It is the intersection between the top-K source domains predicted by the metric and top-K source domains as per CDSA accuracy, for a particular target domain. In other words, it is the number of true positives.
    \item \textbf{Ranking Accuracy (RA):} It is the number of predicted source domains that are \textbf{ranked correctly} by the metric.
\end{itemize}

\noindent Figure \ref{fig:overlap} shows the number of true positives (precision) when K = 5 for each metric over the top 10 domains. The X-axis denotes the domains, whereas the Y-axis in the bar graph indicates the precision achieved by all metrics in each domain. We observe that the highest precision attained is 5, by 4 different metrics. We also observe that all the metrics reach a precision of at least 1. A similar observation is made for the remaining domains as well. Figure \ref{fig:precision} displays the RA values of K = 5 in each metric for the top 10 domains. Here, the highest number of correct source domain rankings attained is 4 by ULM6 (ELMo) for domain D5. \\

\noindent Table \ref{tab:results} shows results for different values of K in terms of precision percentage and normalized RA (NRA) over all domains. Normalized RA is RA scaled between 0 to 1. For example, entries 45.00 and 0.200 indicate that there is 45\% precision with NRA of 0.200 for the top 3 source domains.

These are the values when the metric LM1 (Significant Words Overlap) is used to predict the top 3 source domains for all target domains. 
Best figures for precision and NRA have been shown in bold for all values of K in both labelled as well as unlabelled data metrics. ULM7 (Universal Sentence Encoder) outperforms all other metrics in terms of both precision and NRA for K = 3, 5, and 7. When K = 10, however, ULM6 (ELMo) outperforms ULM7 marginally at the cost of a 0.02 degradation in terms of NRA. For K = 3 and 5, ULM2 (Doc2Vec) has the least precision percentage and NRA, but UML3 (GloVe) and ULM5 (FastText Sentence) take the lowest pedestal for K = 7 and K = 10 respectively,  in terms of precision percentage.

\section{Discussion}
\label{sec:disc}

Table \ref{cross-domain} shows that, if a suitable source domain is not selected, CDSA accuracy takes a hit. The degradation suffered is as high as 23.18\%. This highlights the motivation of these experiments: the choice of a source domain is critical. We also observe that the automative domain (D2) is the best source domain for clothing (D6), both being unrelated domains in terms of the products they discuss. This holds for many other domain pairs, implying that mere intuition is not enough for source domain selection.

From the results, we observe that LM4, which is one of our novel metrics, predicts the best source domain correctly for $D_2$ and $D_4$, which all other metrics fail to do. This is a good point to highlight the fact that this metric captures features missed by other metrics. Also, it gives the best RA for K=3 and 10. Additionally, it offers the advantage of asymmetricity unlike other metrics for labelled data.\\

\noindent For labelled data, we observe that LM2 (Symmetric KL-Divergence) and LM3 (Chameleon Words Similarity) perform better than other metrics. Interestingly, they also perform identically for K = 3 and K = 5 in terms of both precision percentage and NRA. We accredit this observation to the fact that both determine the distance between probabilistic distributions of polar words in domain-pairs.

Amongst the metrics which utilize word embeddings, ULM1 (Word2Vec) outperforms all other metrics for all values of K. We also observe that word embeddings-based metrics perform better than sentence embeddings-based metrics. Although ULM6 and ULM7 outperform every other metric, we would like to make a note that these are computationally intensive models. Therefore, there is a trade-off between the performance and time when a metric is to be chosen for source domain selection. The reported NRA is low for all the values of K across all metrics. We believe that the reason for this is the unavailability of enough data for the metrics to provide a clear distinction among the source domains. If a considerably larger amount of data would be used, the NRA should improve.

We suspect that the use of ELMo and Universal Sentence Encoder to train models for contextualized embeddings on review data in individual domains should improve the precision for ULM6 (ELMo) and ULM7 (Universal Sentence Encoder). However, we cannot say the same for RA as the amount of corpora used for pre-trained models is considerably large. Unfortunately, training models using both these recur a high cost, both computationally and with respect to time, which defeats the very purpose of our work \textit{i.e.,} to pre-determine best source domain for CDSA using non-intensive text similarity-based metrics.

\section{Conclusion and Future Work}
\label{sec:concl}

In this paper, we investigate how text similarity-based metrics facilitate the selection of a suitable source domain for CDSA. Based on a dataset of reviews in 20 domains, our recommendation chart that shows the best source and target domain pairs for CDSA would be useful for deployments of sentiment classifiers for these domains.

In order to compare the benefit of a domain with similarity metrics between the source and target domains, we describe a set of symmetric and asymmetric similarity metrics. These also include two novel metrics to evaluate domain adaptability: namely as LM3 (Chameleon Words Similarity) and LM4 (Entropy Change). These metrics perform at par with the metrics that use previously proposed methods. We observe that, amongst word embedding-based metrics, ULM6 (ELMo) performs the best, and amongst sentence embedding-based metrics, ULM7 (Universal Sentence Encoder) is the clear winner. We discuss various metrics, their results and provide a set of recommendations to the problem of source domain selection for CDSA.

A possible future work is to use a weighted combination of multiple metrics for source domain selection. These similarity metrics may be used to extract suitable data or features for efficient CDSA. Similarity metrics may also be used as features to predict the CDSA performance in terms of accuracy degradation.

\section{Bibliographical References}
\label{main:ref}

\bibliographystyle{lrec}
\bibliography{lrec2020W-xample}

\end{document}